\documentclass{article}
\usepackage[final]{corl_2025}
\usepackage{amssymb}
\usepackage{times}
\usepackage[dvipsnames,table,xcdraw]{xcolor}
\usepackage{xspace}
\PassOptionsToPackage{numbers,sort&compress,square}{natbib} 
\usepackage{multicol}
\usepackage{graphicx}
\usepackage{caption}
\usepackage{tcolorbox}
\usepackage{times}
\usepackage{float}
\usepackage{amsfonts}
\usepackage{amsmath}
\usepackage{listings}
\usepackage{mdframed}
\usepackage{soul}
\usepackage{inconsolata}
\usepackage{tcolorbox}
\usepackage{fancyvrb}
\usepackage{caption}
\usepackage{cleveref}
\usepackage{enumitem}
\usepackage{booktabs}
\usepackage{multirow}

\usepackage{times}
\usepackage{xspace}
\usepackage{wrapfig}
\usepackage{graphicx}

\lstdefinestyle{customstyle}{
    language=TeX, 
    basicstyle=\small\ttfamily,
    keywordstyle=\color{blue},
    commentstyle=\color{green!50!black},
    numbers=left,
    numberstyle=\tiny,
    numbersep=5pt,
    breaklines=true,
    showstringspaces=false,
    frame=none,
}

\sethlcolor{orange!30}

\newmdenv[backgroundcolor=blue!5]{boxedsection}

\lstdefinestyle{LLMQuery}{
  basicstyle=\ttfamily,
  breaklines=true,
  frame=single,
  backgroundcolor=\color{gray!10},
  xleftmargin=0pt,
  columns=fullflexible,
  breakindent=0pt,
  rulecolor=\color{black},
  moredelim=**[is][\textit{}]{\%\%}{\%\%},
  moredelim=**[is][\color{red}\textbf{}\bfseries]{\|\|}{\|\|}
}
\definecolor{ferngreen}{rgb}{0.31, 0.47, 0.26}
\lstdefinestyle{LLMReply}{
  basicstyle=\ttfamily,
  breaklines=true,
  frame=single,
  breakindent=0.2pt,
  backgroundcolor=\color{ferngreen!10},
  xleftmargin=0pt,
  rulecolor=\color{black},
  columns=fullflexible,
}

\lstdefinestyle{customstyle}{
    language=TeX, 
    basicstyle=\small\ttfamily,
    keywordstyle=\color{blue},
    commentstyle=\color{green!50!black},
    numbers=left,
    numberstyle=\tiny,
    numbersep=5pt,
    breaklines=true,
    showstringspaces=false,
    frame=none,
}

\newcommand{\secref}[1]{Section~\ref{#1}}
\renewcommand{\eqref}[1]{Eqn~\ref{#1}}
\newcommand{\figref}[1]{Figure~\ref{#1}}

\newcommand{\tabref}[1]{Table~\ref{#1}}

\newcommand{\appendref}[1]{Appendix~\ref{#1}}

\def\ourapproach{SAVOR}
\def\ournetwork{SAVOR-Net}

\newcommand{\ie}{\textrm{i.e.}}
\newcommand{\eg}{\textrm{e.g.}}

\author{
  Zhanxin Wu$^1$ \quad Bo Ai$^2$ \quad Tom Silver$^1$ \quad Tapomayukh Bhattacharjee$^1$\\
  $^1$Cornell University \quad $^2$UC San Diego
}

\title{SAVOR: Skill Affordance Learning from Visuo-Haptic Perception for Robot-Assisted Bite Acquisition}

\begin{document}
\maketitle

\begin{center}
    \vspace{-1cm}
    \includegraphics[width=0.95\textwidth]{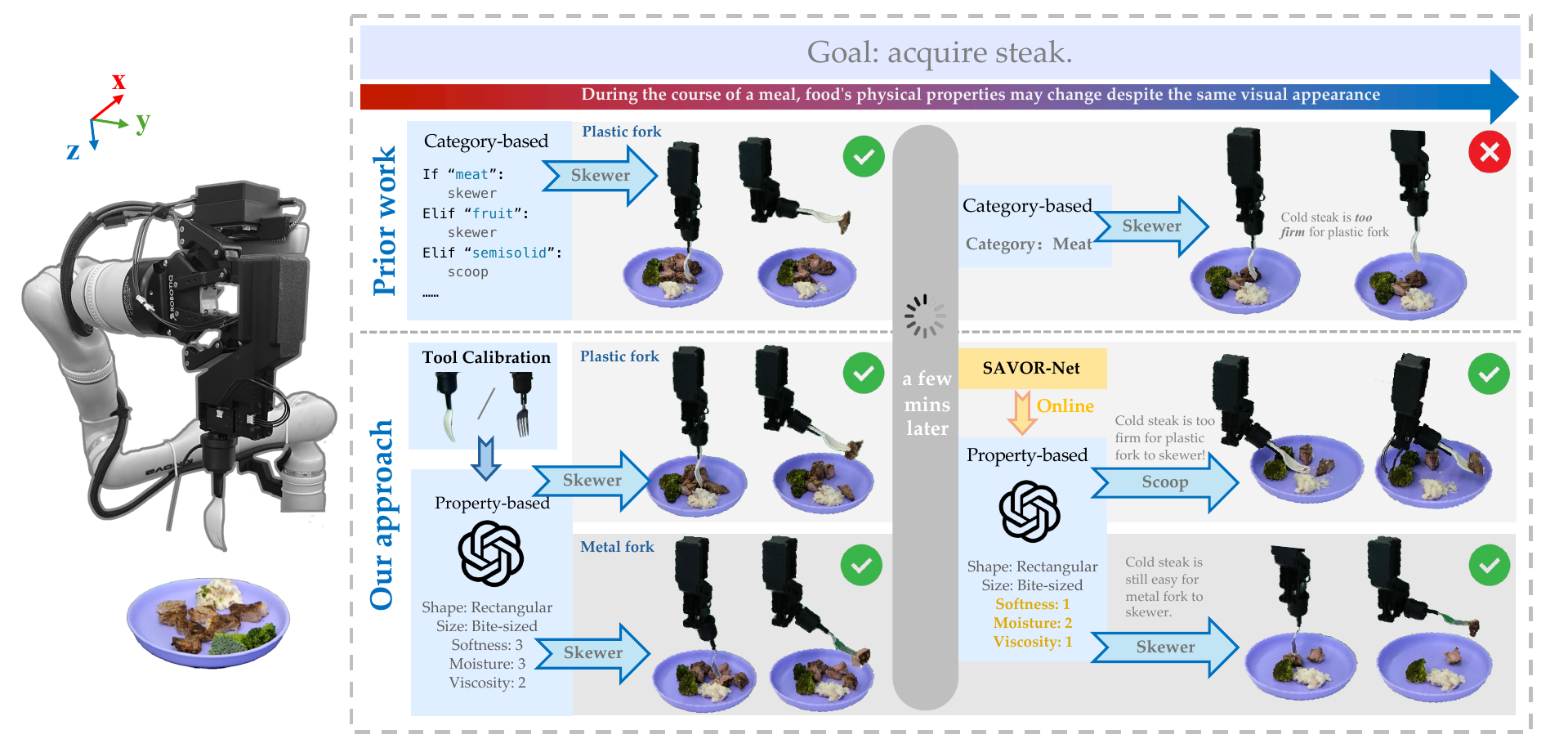} 
    \captionof{figure}{
        We propose \textbf{\ourapproach{}}, a method that combines tool affordances and food affordances to select the appropriate manipulation skill for robust bite acquisition.
    }
    
    \label{fig:teaser}
\end{center}


\begin{abstract}
Robot-assisted feeding requires reliable bite acquisition, a challenging task due to the complex interactions between utensils and food with diverse physical properties. These interactions are further complicated by the temporal variability of food properties—for example, steak becomes firm as it cools even during a meal. To address this, we propose \ourapproach{}, a novel approach for learning skill affordances for bite acquisition—how suitable a manipulation skill (e.g., skewering, scooping) is for a given utensil-food interaction. In our formulation, skill affordances arise from the combination of tool affordances (what a utensil can do) and food affordances (what the food allows). Tool affordances are learned offline through calibration, where different utensils interact with a variety of foods to model their functional capabilities. Food affordances are characterized by physical properties such as softness, moisture, and viscosity, initially inferred through commonsense reasoning using a visually-conditioned language model and then dynamically refined through online visuo-haptic perception using \ournetwork{} during interaction. Our method integrates these offline and online estimates to predict skill affordances in real time, enabling the robot to select the most appropriate skill for each food item. Evaluated on 20 single food items and 10 in-the-wild meals, our approach improves bite acquisition success rate by 13\% over state-of-the-art (SOTA) category-based methods (e.g. use skewer for fruits). These results highlight the importance of modeling interaction-driven skill affordances for generalizable and effective robot-assisted bite acquisition. Website: \textcolor{blue}{https://emprise.cs.cornell.edu/savor}


\end{abstract}



\section{Introduction}
\vspace{-5pt}

Eating is a fundamental human activity, yet millions struggle to feed themselves due to mobility limitations~\cite{WHO2022GlobalHealth}. A robot-assisted feeding system has the potential to help them regain independence and dignity while ensuring their needs are met reliably~\cite{NanavatiHRI23}. Bite acquisition, the process of picking up a food item from a plate or bowl, is a critical step in robot-assisted feeding, but it is highly challenging: 
(i) Food items exhibit diverse and temporally variable physical properties (e.g., rice becomes firm when it cools down, and tofu is fragile and can break without careful manipulation). 
(ii) Physical interaction with food items varies significantly for different utensils. 
To address these challenges, a robot must reason about three types of affordances. First, \textbf{food affordances} describe what the food allows, such as whether the food can be skewered or scooped. Second, \textbf{tool affordances} characterize what a utensil can do, given its functionality. \textbf{Skill affordances} arise from reasoning jointly over food and tool affordances and capture whether a manipulation skill is appropriate, given the food’s physical properties and the tool’s capabilities.

To this end, our key insight is that \textit{a better understanding of skill affordances---achieved by combining calibrated tool affordances and commonsense food affordances that are updated using online information gathering---enhances bite acquisition across diverse food items}.
Take humans as an example. We begin with an understanding of tool affordance, \eg, knowing that a plastic fork may not penetrate firm foods. Upon seeing a food item, we form initial beliefs about its food affordances from visual cues, such as assuming a piece of steak is moderately soft and suitable for skewering. However, these assumptions are not always accurate. During interaction, visual and haptic feedback can reveal unexpected firmness (\eg, a well-done steak), prompting an adjustment of the chosen skill. This process highlights the importance of (i) understanding tool capabilities, (ii) estimating food properties visually, and (iii) refining estimates through visuo-haptic feedback for skill selection. 

\begin{figure}[t]
    \centering
    \includegraphics[width=1\linewidth]{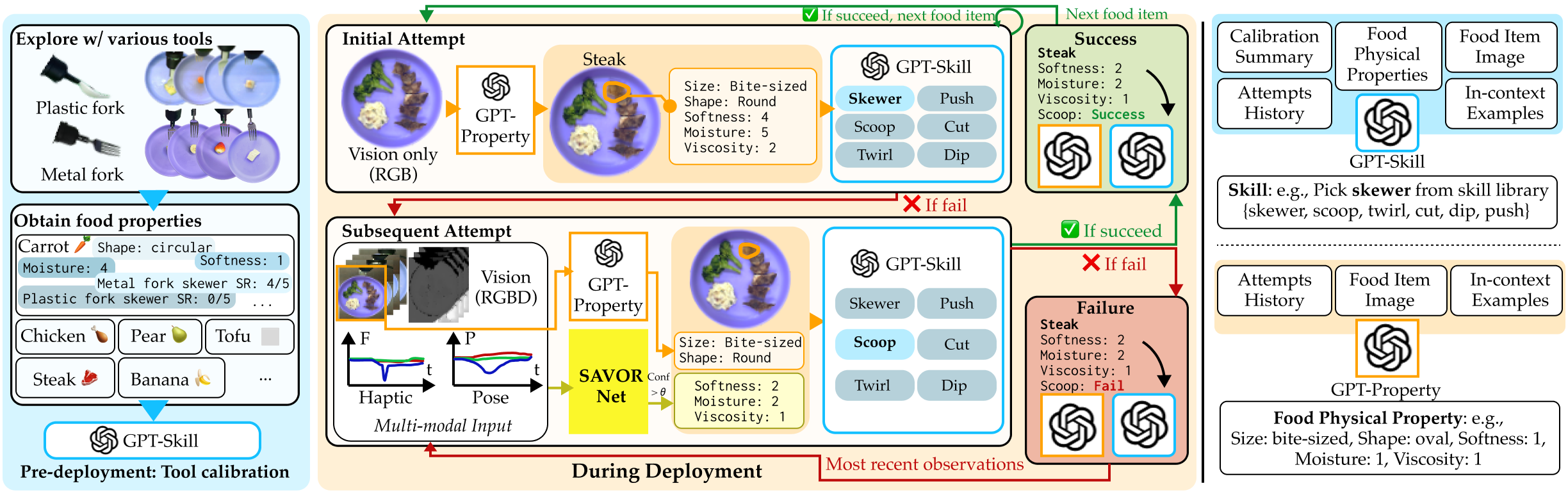}
    \caption{\textbf{\ourapproach{} Framework.} 
    Before deployment, we perform an offline tool calibration to understand tool affordances. During deployment, we first use a visually-conditioned language model to estimate food physical properties and then refine it through online visuo-haptic perception.
    }
    \label{fig:overview} 
    \vspace{-20pt}
\end{figure}

Building on these insights, we propose \ourapproach{}, an approach for learning \textit{skill affordances} from tool affordances and food affordances for bite acquisition (\autoref{fig:overview}).
The system operates in two stages: (i) \textbf{Prior to deployment}, we learn \textit{tool affordances} through offline calibration, where different utensils interact with various foods to evaluate manipulation skills to model tool capabilities. (ii) \textbf{During deployment}, we estimate \textit{food affordances} by inferring food physical properties, including softness, moisture, and viscosity. We initially estimate food properties through commonsense reasoning using a visually-conditioned language model, and dynamically refine them through online visuo-haptic perception with our developed \ournetwork{} during the interaction. Together, combining offline tool affordances and online food affordances guides robust skill selection for bite acquisition.

Overall, our contributions are: (i) an \textbf{algorithm} that learns food affordances based on physical properties, initially informed by commonsense priors and dynamically refined through online visuo-haptic perception; (ii) a \textbf{method} that characterizes the affordances of manipulation skills across diverse food items and tools and uses the learned model for adaptive skill selection; (iii) a \textbf{comprehensive evaluation} on 20 single food items and 10 in-the-wild dishes, showing improved bite acquisition success over state-of-the-art approaches; and 
(iv) a \textbf{dataset} of bite acquisition trials, which is the first to involve varied manipulation skills applied to diverse foods, and provides synchronized visual and haptic signals throughout the trajectory.


\vspace{-5pt}
\section{Related Work}
\vspace{-5pt}
\textbf{Food Manipulation for Robot-Assisted Feeding.}
Feeding~\cite{jenamani2024flair, ha2024repeat, bhattacharjee2019foodhaptics, jenamani2024bitetransfer, nanavati2025lessons} is an important activity of daily living (ADL). 
One crucial step in feeding is bite acquisition, which involves executing various skills, such as skewering~\cite{sundaresan2022learning, gordon2020adaptive, gordon2021leveraging, gordon2023towards, feng2019robot}, twirling~\cite{pmlr-v229-sundaresan23b}, and scooping~\cite{scoop=pmlr-v229-tai23a, grannen2022learning}. 
To compose these skills, VAPORS~\cite{pmlr-v229-sundaresan23b} focuses on noodle dishes and employs physics-based simulations for decision-making between twirling and grouping noodles. The closest relevant work, FLAIR~\cite{jenamani2024flair}, proposes to select skills by food categories inferred from only visual observations.
Such a category-based approach overlooks the intricate physical properties of food, often resulting in failures with items of varying softness. To address this, \textsc{\ourapproach{}} selects manipulation skills based on estimated food affordances to enable more adaptive bite acquisition across diverse food items.

\textbf{Integrating Vision and Haptics in Robotics.}
Vision-only manipulation has demonstrated effectiveness in robotic domains such as semantic grasping and deformable object manipulation~\cite{shridhar2022cliport, Tadhg2004foodvision, gordon2020adaptive, he2025learning, tian2025diffusion}. However, contact-rich tasks, such as in-hand manipulation~\cite{qi2023general, suresh2024neuralfeels} and object packing~\cite{ai2024robopack}, benefit significantly from the integration of vision and haptic information. Previous studies~\cite{bhattacharjee2019foodhaptics,gordon2021leveraging, Tapo2019RAL, Yamaguchi2016Cutting, xu2023roboninja, heiden2021disect, zhang2019leveraging, lenz2015deepmpc,MORPHeus2024Peeling, Dong2021Peeling, pancake2024iros} have highlighted the importance of haptic feedback in food manipulation. 
More recently, Sundaresan et al.~\cite{sundaresan2022learning} and Gordon et al.~\cite{gordon2021leveraging} combine a single image with time-series haptics to learn skewering skills, but rely either on haptics alone or static visual input. In contrast, \ourapproach{} focuses on learning food physical properties from time-series observations of both vision and haptics, allowing us to capture dynamic changes, such as deformation and surface texture variations, which enables a more comprehensive understanding of the food physical properties and their temporal variations.

\textbf{Foundation Models in Robot Manipulation.} 
Foundation models~\cite{Bommasani2021on} are widely used for language-conditioned planning in robotics by prompting them with task context, available skills, and agent state information~\cite{AgiaMigimatsuEtAl2023, huang2022inner, wu2023tidybot, wu2023integrating, Singh2023ProgPrompt, driess2023palme, song2023llmplanner, Ichter2022Saycan, wang2023newton, gao2025do, shi2025learning, wu2024open}. 
Similarly, \ourapproach{} leverages a visually-conditioned language model to form initial beliefs about food physical properties from an image. These estimates are then dynamically refined through visuo-haptic perception during interaction.


\vspace{-10pt}
\section{Problem Formulation}
\vspace{-10pt}

We consider the problem of bite acquisition, where a robot equipped with a utensil needs to acquire food items from a plate in a specified sequence. The user specifies the desired food sequence (e.g., via natural language~\cite{jenamani2024flair}), and the robot executes one or more actions per item to acquire it. Upon a successful attempt, the item is removed from the tool, and the robot proceeds to the next. The objective is to maximize the number of food items acquired within a limited number of attempts.

We formulate this as a Partially Observable Markov Decision Process (POMDP)~\cite{Kaelbling1998planning}, defined by the tuple $(\mathcal{S}, \mathcal{A}, \mathcal{O}_0, \mathcal{O}, \mathcal{T}, \mathcal{Z}, R, L)$. The state $s \in \mathcal{S}$ comprises the robot end-effector pose, the target food item, and the positions and physical properties of all food items on the plate. Actions $a \in \mathcal{A}$ are discrete skills selected from a predefined library: \texttt{\{push, cut, skewer, dip, scoop, twirl\}}~\cite{jenamani2024flair}. The initial observation $o_0 \in \mathcal{O}_0$ consists of a single RGB-D image $I_0 \in \mathbb{R}^{W \times H \times 4}$ from a wrist-mounted camera. Subsequent observations $o \in \mathcal{O}$ include a time series of RGB-D images $I \in \mathbb{R}^{T \times W \times H \times 4}$, force-torque readings $F \in \mathbb{R}^{T \times 6}$, and end-effector poses $P \in \mathbb{R}^{T \times 6}$. The length of the time series $T$ varies depending on the execution of the skill. The target food item and its position are assumed to be observable. The transition model $\mathcal{T}$ and observation model $\mathcal{Z}$ are unknown. The reward function $R : \mathcal{S} \to \mathbb{R}$ returns 1 for successful acquisition of the target item, and 0 otherwise. The time horizon $L \in \mathbb{Z}$ is finite.

Importantly, the physical properties of food are not directly observed, but we hypothesize that the visuo-haptic observations $I$ and $F$ can be used to infer these properties towards completing an estimate of the partially-observable state $s$. We consider five physical properties: shape, size, softness, moisture, and viscosity. We assume that shape and size can be estimated from vision alone, while softness, moisture, and viscosity require integrating vision and haptic feedback. Shape and size are represented as categorical descriptors in natural language (e.g., ``bite-sized,’’ ``round’’), and the remaining properties are scalar values from 1 to 5, similar to a 5-point Likert scale. For example, a softness score of 1 indicates very hard, while a score of 5 indicates very soft. 
We hypothesize that the visuo-haptic observations $(I, F)$ provide sufficient information to infer these physical properties. 
The robot should use its estimates of these latent physical properties to guide skill selection and maximize bite acquisition success.

\vspace{-5pt}
\section{\ourapproach{}} \label{sec:methods}
\vspace{-5pt}


Robust bite acquisition requires reasoning over both food affordances and tool affordances.
To estimate tool affordances, we begin with an offline calibration phase, where the robot executes various skills on diverse food items (\secref{sec:Calibration}). The resulting calibration dataset serves as an implicit representation of the tool’s affordances. In parallel, we offline train \ournetwork{}, a neural network that predicts physical food properties from raw visuo-haptic observations (\secref{sec:NetworkTraining}). At test time, given a new plate of food, we first initialize food physical property estimates using a visually-conditioned language model (VLM), and refine them online with \ournetwork{} (\secref{sec:StateEstimation}). For action selection, we prompt a VLM-based planner with the estimated food properties and the tool calibration dataset to generate a skill sequence (\secref{sec:Planning}). We detail each component below.

\vspace{-5pt}
\subsection{Pre-Deployment: Offline Tool Calibration}
\vspace{-5pt}
\label{sec:Calibration}

The goal of tool calibration is to provide the robot with contextual knowledge of the utensil's affordances. 
We capture tool affordances implicitly by collecting a small offline calibration dataset consisting of randomly sampled skill executions on diverse food items, annotated by humans  with utensil type, food type, physical properties, and execution outcomes. For example:

\begin{tcolorbox}[colback=gray!10, colframe=gray!40, boxrule=0.1pt, sharp corners, fontupper=\small, left=0pt, width=\columnwidth]
\begin{verbatim}
The robot arm interacts with various food items using a plastic fork. We summarize 
the history as follows:
Food Item: Nuts, Shape: Oval, Size: Bite-sized, Softness: 1,  Moisture: 1, Viscosity: 1
Skill with Success Rate: Skewer 0/5, Scoop 3/5, Cut 0/5, Push 5/5, Dip 5/5
\end{verbatim}
\end{tcolorbox}

This calibration dataset serves as an implicit representation of tool affordances, particularly the success rates of various skills with a specific tool, expressed in natural language and later used as input for VLM-based planning (\secref{sec:Planning}).  

\vspace{-5pt}
\subsection{Pre-Deployment: Offline Training \ournetwork{}}
\vspace{-5pt}
\label{sec:NetworkTraining}

\begin{figure}[t!]
     \centering
     \includegraphics[width=1\linewidth]{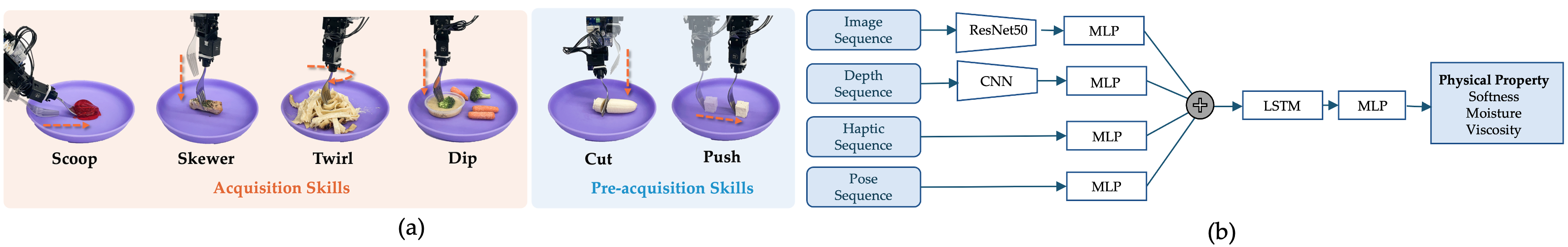}
     \caption{\textbf{(a) Skill library for bite acquisition.}
     \textbf{(b) \ournetwork{} model architecture.}
     }
     \label{fig:architecture}
     \label{fig:skill}
     \vspace{-15pt}
 \end{figure}

While still offline, we next train \ournetwork{}, a neural network that predicts physical food properties (\ie, softness, moisture, and viscosity) from visuo-haptic observations.
We discretize food property values into $C=5$ levels (e.g., $\texttt{Viscosity} \in \{1, 2, \dots, 5\}$).
The input to \ournetwork{} at time $t$ includes all three time-series observations $(I_t, F_t, P_t)$ and the output is $\psi_t \in \mathbb{R}^{3 \times C}$, a vector of log probabilities for each of the $3$ predicted food properties: softness, moisture, and viscosity. We select these three properties based on prior work~\cite{Gemici2014, ha2024repeat} and insights obtained by querying a VLM about which food physical properties are relevant for the bite acquisition task.
\ournetwork{} uses separate encoders for each of the time series and further splits the RGB-D inputs into RGB and depth for separate encoding.
Each encoder outputs a vector in $\mathbb{R}^{128}$.
The four vectors are concatenated into a unified multimodal representation and then passed to an LSTM with 2 layers and a hidden size of 512.
A three-layer MLP takes output from the LSTM and produces the final output $\psi_t$. We present the structure
of the model in \figref{fig:architecture}. 

We pretrain \ournetwork{} on an existing dataset~\cite{DVN/C8SI1D_2022}, which contains 400 examples of human skewering various food items, and then fine-tune it on our own dataset, consisting of 300 examples of a robot performing skewering, pushing, twirling, cutting, dipping, and scooping across 20 food items. More details about the model and the training process are available in \appendref{appendix:training}.

\vspace{-5pt}
\subsection{During Deployment: State Estimation} \label{sec:StateEstimation}
\vspace{-5pt}
Given a tool calibration dataset and a trained \ournetwork{}, the robot is ready to acquire bites from novel dishes and food items (\figref{fig:overview}). At the first time step, it initializes a state estimate using only visual observations. As the robot attempts to acquire a target item (e.g., \texttt{cutting} then \texttt{skewering} a banana), it collects visuo-haptic data for \ournetwork{} to refine the estimate, which is especially important when the attempt fails. Once the item is acquired, a new target is specified. Although the robot initially lacks physical knowledge of the new item, it can leverage prior interaction history for improved estimation (e.g., banana slices on the same plate likely share physical properties). We now describe the state estimation process in more detail.

\textbf{Initializing State Estimates.} 
Given an initial RGB-D observation $I_0$, we prompt a VLM (GPT-4V~\cite{achiam2023gpt}) to extract semantic labels of food items (e.g., $\texttt{[`potato', 'chicken']}$). We then use GroundingDINO~\cite{liu2023grounding} to obtain segmentation masks for them.
Finally, we prompt VLM again with the masks and in-context examples (\appendref{appendix:prompting}) to estimate physical properties for each item. These estimates are used by a VLM planner to select an appropriate skill (\secref{sec:Planning}).

\textbf{Refining State Estimates with \ournetwork{}.} If the skill (\secref{sec:Planning}) selected based on the initial estimate of food physical properties fails, we use \ournetwork{} to refine the state estimate for the target food item for each timestep $t > 0$.
The time series $(I_t, F_t, P_t)$ in $o_t$ are input to \ournetwork{}, which produces log probability outputs $\psi_t$.
For each predicted property, if its log probability is less than a threshold $\theta_{th}$, the prediction is ignored.
Otherwise, the estimate is updated. This process yields a refined $\hat{s}_t$, where only the target item's properties are potentially modified from $\hat{s}_{t-1}$. 

\textbf{Generating State Estimates for New Targets.} When the target item changes at timestep $t > 0$ and no interaction data is available, we initialize the estimate using prior outcomes on similar items. We run the same detection and segmentation pipeline, then prompt a VLM (\appendref{appendix:prompting}) with: (i) segmented images, (ii) past attempt summaries, (iii) the previous estimate $\hat{s}_{t-1}$, and (iv) in-context examples. The VLM predicts $\hat{s}_t$ by matching the new item’s visual features to similar examples. This estimate is later refined through interaction, using \ournetwork{} as mentioned above.

\vspace{-5pt}
\subsection{During Deployment: Planning}
\vspace{-5pt}
\label{sec:Planning}
Given the tool calibration dataset and the estimated food state $\hat{s}_t$, we query a VLM (GPT-4V~\cite{achiam2023gpt}) to select a skill from the skill library. Our prompt for the VLM includes (i) the calibration dataset; (ii) a brief description of each skill;  (iii) a natural-language history of acquisition attempts; (iv) the segmented food item image; and (v) the estimated physical properties of the target food item.
Below, we present an example prompt. 
The full prompting strategy is detailed in \appendref{appendix:prompting}.

\begin{tcolorbox}[colback=gray!10, colframe=gray!40, boxrule=0.1pt, sharp corners, fontupper=\small, left=0pt,]
\footnotesize
\begin{verbatim}
[Calibration Summary] [Skill descriptions] [History of acquisition attempts]
This is a food item: Mashed Potatoes. <Image>
The robot uses a plastic fork to try picking up the food.
The food physical properties, which range from 1 to 5, are as follows:
Shape: Circular, Size: bite-sized, Softness: 4, Moisture: 3, Viscosity: 3
Please select an action from ['skewer, 'scoop', 'twirl', 'dip'] to pick up the food item or 
select one action from ['cut', 'push'] to manipulate items to facilitate subsequent 
acquisition. Always follow the format:  Reasoning: <your reason>. Answer: <your answer>.
\end{verbatim}
\end{tcolorbox}




For the low-level skills, we use the skill library from FLAIR~\cite{jenamani2024flair}, which includes four acquisition skills (\texttt{skewer}, \texttt{twirl}, \texttt{scoop}, \texttt{dip}) and two pre-acquisition skills (\texttt{cut}, \texttt{push}), illustrated in Figure~\ref{fig:skill}. 
Given these low-level skills, \ourapproach{} enables effective skill selection for each food item, despite variations in physical properties and the partial observability present during deployment.



\vspace{-5pt}
\section{Experiments}
\vspace{-5pt}
\label{sec:experiments}

In this section, we aim to answer the following questions: \textbf{Q1.} Does tool calibration contribute to effective skill selection? \textbf{Q2.} How does visuo-haptic feedback help estimate the physical properties of food items? \textbf{Q3.} How effective is our overall approach compared to prior methods? \textbf{Q4.} How well does \ourapproach{} generalize to unseen food items?


\begin{figure*}[t]
    \centering
    \includegraphics[width=1\textwidth]{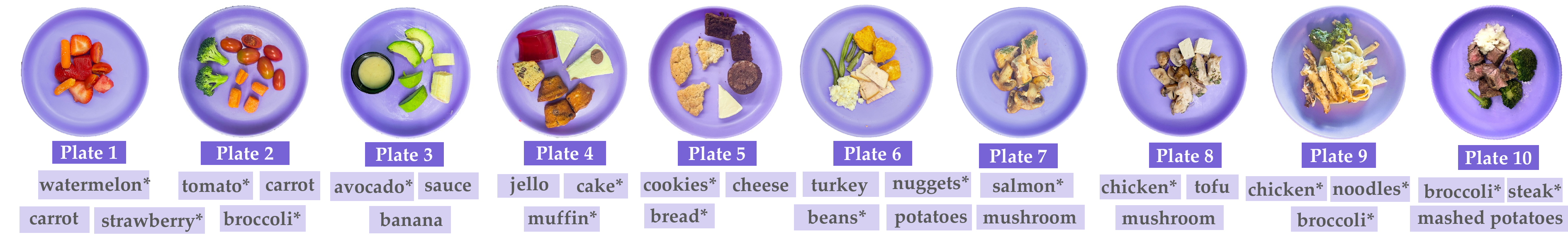}
    \caption{\textbf{Experimental setup: 10 in-the-wild dishes.}   
    $^\ast$ denotes food items unseen during training.
    }  
    \vspace{-10pt}
    \label{fig:wild_plate}
\end{figure*}

\begin{table*}[t]
\centering
\caption{\textbf{Quantitative results on bite acquisition.} 
We evaluate our approach on 10 different dishes (\autoref{fig:wild_plate}) and report both Attempt Efficiency (AE) and Success Rate (SR), averaged across plates.}
\scriptsize
\begin{tabular}{c | c c c c c c c}
\toprule
\multirow{2}{*}{\textbf{Plate}} 
  & \multicolumn{7}{c}{\textbf{\#Items Acquired / \#Total Attempts}} \\ \cline{2-8}
  & \textbf{\ourapproach{}} & \textbf{Haptic-only} & \textbf{Vision-only} 
  & \textbf{FLAIR}  & \textbf{VLM} & \textbf{SayCan} & \textbf{End2End} \\ \hline
1  & \textbf{10/15} & 9/17 & 6/26 & 9/18 & 9/19 & 9/22 & 9/15 \\
2  &  \textbf{7/13} & 4/17 & 4/18 & 3/15 & 4/17 & 5/13 & 4/13 \\
3  & \textbf{7/11} & \textbf{7/11} & 6/14 & 5/11 & \textbf{7/11} & 5/15 & 6/11 \\
4  & 6/12 & 6/13 & 6/13 & 5/15 & 5/14 & \textbf{7/10} & 6/12  \\
5  & 6/13 & 6/15 & 5/21 & 5/14 & \textbf{ 7/12} & 5/14 & 5/17 \\
6  & \textbf{10/18} & 8/20 & 7/29 & 8/23 & 8/23 &8/23 & 8/16\\
7  & 5/17 & 6/17 & 4/19 & 5/17 & 6/14  & 4/15 & \textbf{6/12}\\
8  & 5/9 & 4/12 & 4/16 & 4/12 & 5/10 & 5/12 & \textbf{6/9}\\
9  & \textbf{7/10} & 6/12 & 5/17 & 7/11 &\textbf{ 7/10}  & 5/18 & 4/17\\
10 & 6/16 & 5/18 & 5/19 & \textbf{7/16} & 6/16 & 5/18 & 5/16 \\ 
\hline
\multicolumn{1}{c|}{Average AE (\%)} & \textbf{51.5 $\pm$ 12.7 } & 40.1 $\pm$ 12.2  & 27.1 $\pm$ 8.8  & 38.2 $\pm$ 12.2  & 43.8 $\pm$ 14.3  & 36.2 $\pm$ 12.6 & 42.7 $\pm$ 14.7 \\
\multicolumn{1}{c|}{SR3 (\%)}  & \textbf{87.3 $\pm$ 10.0} & 77.2 $\pm$ 13.0  & 65.8 $\pm$ 11.2  & 73.4 $\pm$ 15.2  & 81.0 $\pm$ 13.1  & 73.4 $\pm$ 12.4 & 81.8 $\pm$ 14.7 \\
\bottomrule
\end{tabular}
\label{tab:results}
 \vspace{-15pt}
\end{table*}

\vspace{-5pt}
\subsection{Setup}
\vspace{-5pt}
\textbf{Evaluation Scenarios.} We evaluate our approach on a diverse set of food items, including 20 individual food items and 10 in-the-wild dishes (\figref{fig:wild_plate}). The dishes comprise three categories: (i) fruits and appetizers, (ii) grocery store and restaurant meals, and (iii) homemade dishes.

\textbf{Baselines.} 
We evaluate three ablations of our approach and compare against four existing methods. The ablations are: (i) \textbf{\ourapproach{} w/o calibration}: In this ablation, we do not provide calibration information to the VLM planner. (ii) \textbf{Vision-only \ourapproach{}} and (iii)\textbf{ Haptic-only \ourapproach{}}: These two ablations differ from ours in that it has no haptic/visual perception respectively. We also compare against the following baselines: (i) \textbf{FLAIR}~\cite{jenamani2024flair}: This state-of-the-art approach selects bite acquisition action based on food category. (ii) \textbf{VLM w/o history}: This method queries a VLM to predict the physical properties of the food from a single post-contact RGB image and subsequently selects a skill based on the estimated properties. (iii) \textbf{SayCan}~\cite{Ichter2022Saycan}: This approach relies on a pre-contact image to estimate instruction relevance and success likelihood using a value function, which we train on the same \ourapproach{} dataset. (iv) \textbf{End2End}: This method takes the same input as \ournetwork{} but directly predicts bite acquisition actions for execution. 
More details in \appendref{appendix:baselinesdetails}.


\begin{figure}[t]
    \centering
    \includegraphics[width=1\linewidth]{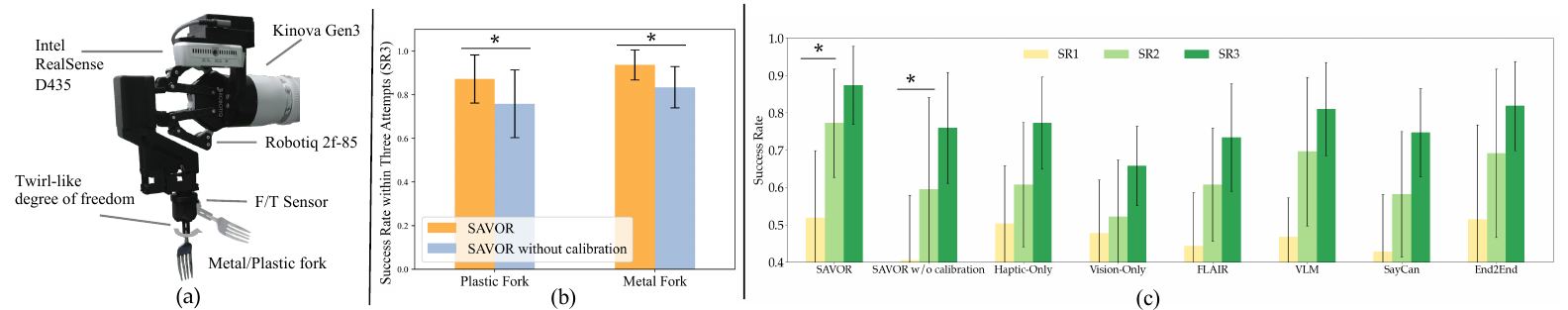}
    \caption{(a) System setup. (b) SR3 using a plastic/metal fork for 10 in-the-wild dishes. (c) SR\{1,2,3\} for 10 dishes. $*$ indicates statistically significant differences ($p < 0.05$).}
    \label{fig:ablation_calibration}
    \label{fig:hardware} 
    \label{fig:example_each_property}
    \label{fig:attemptnum}
    \vspace{-0.5cm}
\end{figure}

\textbf{Evaluation Metrics.} We evaluate each method using \textit{Success Rate}. A trial is considered successful if the utensil picks up the food item and retains it for at least 5 seconds. Failed attempts leave the item on the plate, allowing up to three re-attempts before it is manually removed and recorded as a failure. To capture the efficiency of acquisition, we define $\text{Average Attempt Efficiency (average AE)} = \frac{\# \text{Items Acquired}}{\# \text{Total Attempts}}$ to quantify the proportion
of successful acquisitions relative to total attempts across all plates. To further quantify overall performance, we define \textbf{SR1} as the proportion of items acquired within the first attempt in a meal, i.e., SR1 = $\frac{\# \text{Items Acquired}}{\# \text{Total Items}}$ within one attempt. Similarly, \textbf{SR2} and \textbf{SR3} represent the proportion of items acquired within two and three attempts, respectively.


\textbf{Hardware.} \ourapproach{} is implemented on a Kinova Gen3 robot arm equipped with a motorized feeding utensil at the end-effector~\cite{jenamani2024flair} (\figref{fig:hardware}a). The utensil includes a metal/plastic fork attachment with two degrees of freedom, fork orientation and tilt angle.
We use an Intel RealSense D435 camera mounted on the arm wrist for visual perception and a Nano25 F/T sensor for haptic perception.

 \begin{figure*}[t]
     \centering
     \includegraphics[width=1\linewidth]{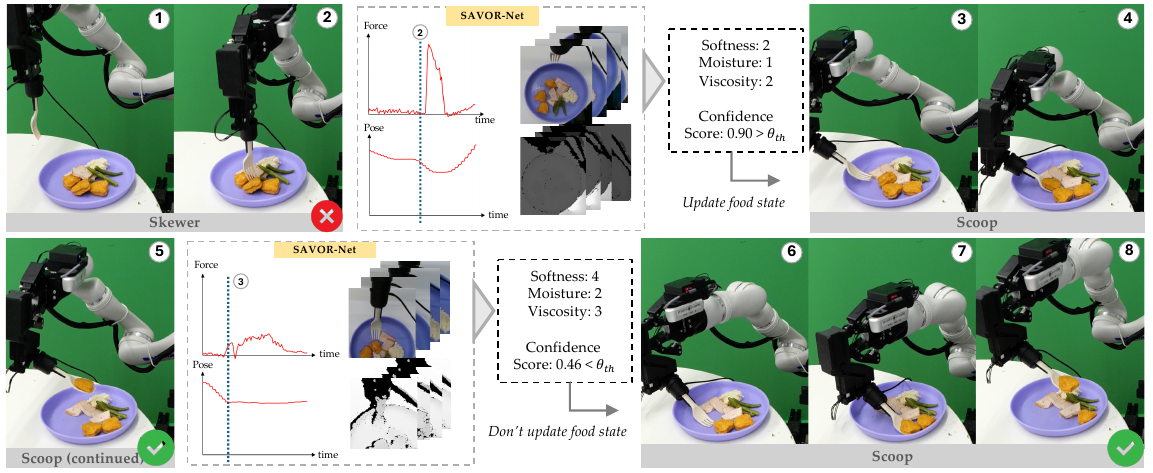}
    \caption{\textbf{Qualitative results on bite acquisition.} 
  The robot first attempts to skewer the food based on its initial property estimate but fails (step 2). Vision and haptic data from this attempt are processed by \ournetwork{}, refining the estimate with high confidence. A VLM planner then selects the scoop skill based on this update (step 3). After executing the scoop (step 4), \ournetwork{} outputs a low-confidence estimate, so the properties are not updated in step 5, and the system proceeds directly to skill selection and execution (steps 6–8).
    }
     \label{fig:example_chicken_nuggets}
     \vspace{-10pt}
 \end{figure*}

  \begin{figure*}[t]
     \centering
     \includegraphics[width=1\linewidth]{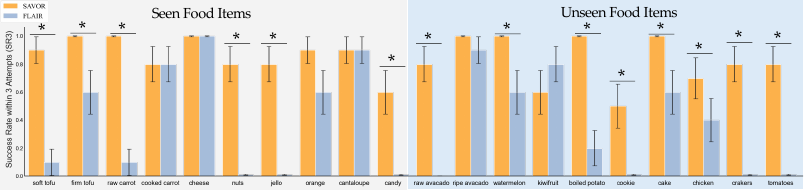}
    \caption{\textbf{Generalization performance on seen and unseen food items.} We compare \ourapproach{} and SOTA FLAIR across 20 food items. We show SR3 on 10 episodes per food item, where each episode allows up to 3 attempts. Asterisks ($*$) indicate statistically significant differences ($p < 0.05$).}
     \label{fig:singlefooditem} 
     \vspace{-21pt}
 \end{figure*}

\vspace{-5pt}
\subsection{Results and Analysis}
\vspace{-5pt}

\textbf{Evaluating the Contribution of Tool Calibration. }
We evaluate the effectiveness of offline tool calibration using two utensils: a plastic fork and a metal fork. As shown in Figure~\ref{fig:ablation_calibration}b and Figure~\ref{fig:ablation_calibration}c, tool calibration significantly improves success rates ($p < 0.05$), leading to an 18\% increase in SR2 and a 13\% increase in SR3.
Without tool calibration, the system selects inappropriate skills, such as skewering a firm steak with a plastic fork. Similarly, without calibration with a metal fork, the system tries to skewer soft, high-moisture foods such as tofu instead of scooping them. Without calibration, \ourapproach{} overuses skewering regardless of the context, leading to frequent failures. Tool calibration helps the system align skill selection with the physical interaction between the utensil and the food, resulting in more reliable bite acquisition.

\textbf{Evaluating the Contribution of Visuo-Haptic Perception.}
We next evaluate how perception modalities affect performance by comparing \ourapproach{} with its vision-only and haptics-only variants (Table~\ref{tab:results}). \ourapproach{} achieves the highest success rate, confirming that combining vision and haptics improves food property estimation and significantly improves performance from SR1 to SR2 in \autoref{fig:attemptnum}c ($p < 0.05$).
The vision-only variant performs the worst (27.1\% average AE), struggling with fine-grained utensil–food interactions and often misassigning visually similar items with inaccurate physical properties. For example, strawberries, watermelon, and carrots on Plate 1 appear similar in color, but differ significantly in physical properties. The model frequently overestimates softness for firm items like carrots, leading to repeated skill selection failures. The haptics-only variant performs better (40.1\% average AE) but still falls short of \ourapproach{}. The haptics-only variant often fails when food items share similar force signatures, such as tofu and nuts. In these cases, small changes in end-effector pose lead to rapid force increases—either due to skewering through soft items and contacting the plate, or due to deformation of the utensil against firm items. Without visual input, these ambiguities lead to misinterpretation of food properties. By integrating vision and haptics, \ourapproach{} resolves ambiguities, achieving a 51.5\% average AE and 87.3\% SR3. \autoref{fig:attemptnum}c demonstrates that \ourapproach{} effectively learns from interaction during the initial attempt and adapts its skill selection dynamically, resulting in the greatest improvement in subsequent attempts. These results highlight the critical role of time-series visuo-haptic perception in estimating food affordances.

\textbf{Evaluating the Overall Framework.}
To evaluate our full framework, we compare \ourapproach{} with the SOTA method (FLAIR), a vision-only variant using a VLM, SayCan and the end2end method. \ourapproach{} achieves the highest average AE (51.5\%) and 87.3\% SR3 across 10 diverse plates (Table~\ref{tab:results}). 

FLAIR selects skills based on predefined food categories, achieving 38.2\% average AE and 73.4\% SR3. It performs well on stereotypical items (e.g., banana), but fails on atypical foods such as raw avocado, where category-based reasoning leads to inappropriate actions (e.g., skewering instead of scooping). VLM improves upon FLAIR, achieving 43.8\% average AE and 81.0\% SR3 by inferring food properties from visual input. However, it struggles with visually ambiguous or dynamically changing items, such as misclassifying avocado or failing to detect firmness changes in cooling chicken nuggets. Similarly, SayCan achieves only 36.2\% average AE and 73.4\% SR3, often oscillating between skewer and scoop skills. Without compact physical property representations, the end-to-end policy generalizes poorly, reaching 42.7\% average AE and 81.8\% SR3. 

While all baselines show improvements from SR1 to SR3 in \autoref{fig:attemptnum}c, these gains largely result from occasional success through repeated trials rather than informed adaptation. Among all methods, only \ourapproach{} and its variant without calibration achieve significant improvement from SR1 to SR2 ($p < 0.05$). As seen in \autoref{fig:attemptnum}c, \ourapproach{} acquires 53.2\% food items with its initial attempt (SR1). By inferring food properties during the initial attempt, the system adapts its skill selection and successfully picks up 77.2\% of the food items with two attempts (SR2), further achieving 87.3\% SR3. It dynamically adjusts skills (e.g., switching from skewering to scooping as nuggets firm up; \figref{fig:example_chicken_nuggets}) and generalizes across similar items on the same plate. These results demonstrate the effectiveness of \ourapproach{} for adaptive bite acquisition.

\textbf{Evaluating Generalization to Unseen Food Items.}
We evaluate \ourapproach{} on 20 single food items, including 10 that are unseen during training. As shown in Figure~\ref{fig:singlefooditem}, \ourapproach{} achieves higher SR3 ($p < 0.05$) than the SOTA baseline FLAIR in 70\% of food items, including raw avocado, cookie, and chicken. These results highlight the advantage of explicitly estimating physical properties over relying on fixed food categories. \ourapproach{} achieves the highest overall SR3 and maintains comparable performance on both seen and unseen items, demonstrating strong generalization.

\textbf{Discussion.}
Our work highlights the importance of grounding skill affordance learning in physical interaction by combining tool and food affordances for effective bite acquisition. This integration allows the system to move beyond state-of-the-art category-based methods and instead make context-sensitive decisions based on how food interacts with a given tool. A key insight from our study is that affordances are not static properties, but rather emerge dynamically through interaction—reflecting how the tool and food jointly shape which skills are applicable. Moreover, we find that visuo-haptic sensing plays a critical role in understanding temporal variations of food physical properties and resolving ambiguity, particularly in cases where visual or haptic cues alone may be misleading. These findings underscore the importance of selecting manipulation skills that are physically grounded in food–utensil interaction for effective bite acquisition.

\section*{Limitations} \label{sec:limitations}

Our approach effectively leverages visuo-haptic sensing to estimate food physical properties and tool calibration to assess tool capabilities for adaptive bite acquisition.
\ourapproach{} currently treats each food item as a whole when estimating its properties. However, a single food item may contain regions with differing physical properties. For instance, foods like broccoli or cauliflower can vary in softness across different parts due to their fibrous stems. While treating each food item as a whole has proven effective, incorporating a more fine-grained food representation could further improve bite acquisition success.
Second, while our system already performs well with open-loop skill execution, it could further benefit from a closed-loop low-level policy to enhance real-time adaptability. 
Finally, slippage during manipulation can introduce noise in haptic signals, leading to inaccurate physical property estimates, which could be mitigated by detecting contact onset, loss, or lateral motion. These directions present promising avenues for future work.



\acknowledgments{This work was partly funded by National Science Foundation IIS \#2132846, and CAREER \#2238792. We thank Ruolin Ye and other members of the EmPRISE Lab for their help with the figures in the manuscript.}


\clearpage
\bibliography{example}  

\clearpage
\appendix
\section*{Appendix}

\section{Additional Results}

\subsection{Timing and Latency Analysis}
We present a timing analysis in Table~\ref{table:latency}, highlighting the processes involving VLM queries. The query for initial state estimation takes about 3.7 seconds to predict the food item’s physical properties from an image. This step occurs only once per item. In subsequent attempts, \ournetwork{} updates the physical properties in 0.2 seconds. As shown in Table~\ref{table:latency}, the major source of latency stems from querying the VLM, and we envision that ongoing work on efficient VLMs holds promise for reducing query timing.

\begin{table}[h!]
\centering
\small
\captionsetup{}
\caption{\textbf{Timing of each component in \ourapproach{}.} $\ast$ indicates processes that include VLM queries.}
\begin{tabular}{c|c|cc|c|c}
\multirow{2}{*}{} & Perception       & \multicolumn{2}{c|}{State Estimation} & Planning        & Control         \\ \cline{2-6} 
 & Object Detection$\ast$ & Initial Attempt$\ast$  & Subsequent Attempt & Skill Selection$\ast$ & Skill Execution \\ \hline
Time (s)& 2.59 $\pm$ 0.32  & 3.69 $\pm$ 0.82      & 0.21 $\pm$ 0.01     & 3.58 $\pm$ 0.74  & 8.54 $\pm$ 1.21        
\end{tabular}
\label{table:latency}
\end{table}

\subsection{Ablation Study on Tool Calibration}
We provide detailed results of the ablation study on the calibration process for the 10 in-the-wild dishes (\tabref{tab:ablation-calibration}). Compared to the uncalibrated baseline, tool calibration significantly improves performance for both the plastic and metal forks. Specifically, with calibration, the plastic fork (PF) achieves a 51.5\% average attempt efficiency and 87.3\% SR3, compared to only 38.7\% and 75.9\% without calibration (PF-wo). Similarly, the metal fork (MF) benefits from calibration, improving from 83.5\% SR3 to 93.7\%. These results demonstrate that understanding tool capabilities through calibration helps the planner avoid infeasible actions, such as skewering tofu with a metal fork (Plate 8) or skewering firm steak with a plastic fork (Plate 10), thereby improving skill selection and acquisition success.

\begin{table*}[h!]
\centering
\tiny
\caption{\textbf{Ablation study on calibration.} Success rates of bite acquisition across 10 in-the-wild dishes, comparing the impact of calibration. PF: Plastic fork; PF-wo: Plastic fork without calibration; MF: Metal fork; MF-wo: Metal fork without calibration. Asterisks ($\ast$) indicates unseen food items.}
\begin{tabular}{c|cccc}
\toprule
\multirow{2}{*}{\textbf{Plate}} 
  & \multicolumn{4}{c}{\textbf{\#Items Acquired / \#Total Attempts}} \\ \cline{2-5}           & \textbf{PF} & \textbf{PF-wo} & \textbf{MF} & \textbf{MF-wo}  \\ \hline
1     & 10/15   & 9/19      & 11/15    & 11/15       \\  
2     &         7/13        &    4/17            &   6/15 &    5/17  \\
3     &          7/11          &       5/15       &      7/10           &     7/9  \\  
4    &    6/13          &     6/13         &     7/11 &    7/11 \\
5         &       6/13          &      7/12       &        7/12      &     7/12  \\ 
6    &         10/18        &      8/21        &         10/17     &    10/18  \\
7    &       5/17          &   6/17           &    7/17 &   7/15  \\
8       &         5/9        &       3/12           & 6/9 &    5/11    \\
9      &     7/10            &      7/10           &      6/12  & 6/12  \\
10         &       6/16          &      5/19    &      8/13     & 7/13 \\  \hline
\multicolumn{1}{c|}{Average AE (\%)}   &        51.5\%               &    38.7\%    &    56.1\%      & 54.1\%  \\
\multicolumn{1}{c|}{SR3 (\%)}   &        87.3\%        &      75.9\%         &    93.7\%    &      83.5\% \\
\bottomrule
\end{tabular}
\label{tab:ablation-calibration}
\end{table*}

\subsection{Crumbly and Soft Foods}
We evaluate our approach on foods that are particularly soft or crumbly, such as tofu and crackers. For crackers, our system selects scooping in 80\% of trials and attempts skewering in 20\% in the initial attempt. After a skewering attempt, \ourapproach{} identifies low softness, logs the failure in the attempt history, and switches to scooping. For tofu, we test variants from extra soft to super firm. The system initially selects skewering for all tofu types but switches to scooping for extra soft tofu. We achieve a 70\% average attempt efficiency for crackers and a 95\% average attempt efficiency for tofu. These results suggest that our method can adapt to diverse food items when informed by interaction feedback.

\begin{figure*}[t]
     \centering
     \includegraphics[width=0.8\linewidth]{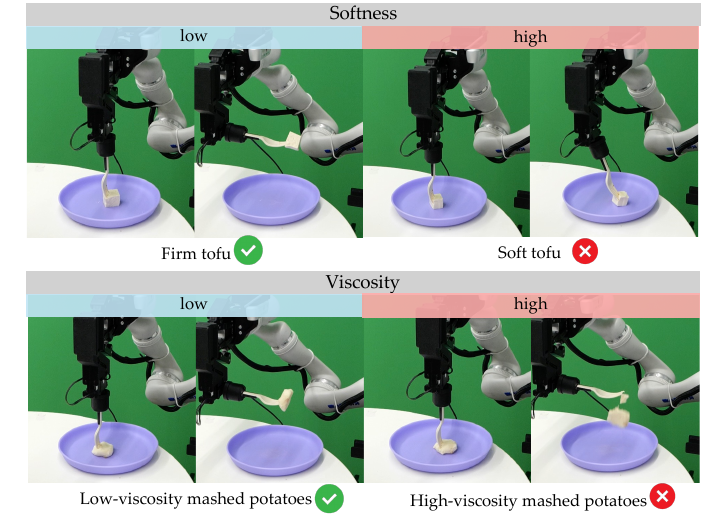}
    \caption{\textbf{Effect of food physical properties on utensil interactions.} The robot skewers food items of varying \textit{softness} (top) and \textit{viscosity} (bottom). Soft tofu and low-viscosity mashed potatoes are successfully acquired, while firm tofu and high-viscosity mashed potatoes lead to failure, illustrating the challenges of bite acquisition.}
     \label{fig:foodpropexample} 
 \end{figure*}
\subsection{VLMs Failures}
We use a single VLM (GPT-4V) for both state estimation and skill planning, instantiated separately as GPT-Property and GPT-Skill. While the VLM itself is static, we augment its context after each interaction with the latest observations. After every interaction, GPT-Property can update food properties. GPT-Skill can also adjust skill choice because some of its inputs are from GPT-Property.

Most VLM-related failures stem from incorrect initial food property estimates by GPT-Property, when it infers physical properties from an RGB image before interaction. These vision-only priors can lead to misidentifications (e.g., confusing carrots with tomatoes) or overlook intra-category variations (e.g., assuming all avocados are soft). This also affects GPT-Skill as GPT-Skill uses inputs from GPT-Property. SR1=$\frac{\# \text{Items Acquired}}{\# \text{Total Items}}$ within 1 attempt ($\sim$50\%) highlights the challenge of using vision priors alone to estimate food affordances, often leading to failures in first attempts. These failures lead to more attempts and thus impact the overall average attempt efficiency) in Table 1, though they are later corrected using online visuo-haptic updates (SR2: 77.2\%, SR3: 87.3\%).

\subsection{Open-loop Nature of Skill Execution}
As mentioned in our limitations section, though our system performs well
with open-loop execution, we acknowledge that it could further benefit from a closed-loop policy, and we plan to address this in future work. However, we conduct further analysis of the experiments and find that the need for more closed-loop skills occurs in only specific cases such as slippage during picking up oily surfaces of salmon and mushrooms, which account for only 7.93\% of trials. Note, despite slippage, re-attempts can potentially pick up the food item.

\subsection{Multi-food Interactions on Cluttered Plates} Our study addresses multi-food interactions on cluttered plates. 8 out of 10 dishes contain overlapping food items, where interactions with one food item affect another food item. In such challenging cases, \ournetwork{} often gives low confidence for push actions, but when the food is pushed toward a cluster of rigid items, it provides high confidence scores and meaningful property estimates.

\section{Implementation Details}

\subsection{Baselines}
\label{appendix:baselinesdetails}
We provide implementation details for the baselines as follows:

\textbf{(i) SayCan~\cite{Ichter2022Saycan}}: This method selects a skill by combining two scores: the skill’s relevance to the instruction and its predicted likelihood of success. As the original work does not release the value function, we train a value function using our \ourapproach{} dataset. The model takes a single RGB image as input and outputs success probabilities for each skill in our predefined skill library. For relevance estimation, we follow the original SayCan setup and use a vision-language model to compute instruction-skill alignment scores.

\textbf{(ii) End2End}: 
We train an end-to-end model for action selection in bite acquisition as a baseline.
This model takes the same input as \ournetwork{}, which includes vision, haptics, and robot poses, and directly predicts one of the six manipulation skills: skewering, scooping, twirling, pushing, dipping, or cutting. The model is trained on the \ourapproach{} dataset.

\subsection{Data Collection}
We collect data by applying each skill from a predefined skill library to food items. The library includes six manipulation skills: skewering, scooping, twirling, pushing, dipping, and cutting. For each skill, we perform 5 trials per food item, recording synchronized RGB-D images, haptic feedback, and pose data throughout each trajectory.
The food items span a range of physical properties and include: bagel, nuts, mashed potatoes, broccoli, jello, carrot, tofu, pork, orange, cantaloupe, candy, lettuce, avocado, cheese, turkey, noodles, watermelon, banana, and tomatoes, along with variations in their cooking or ripeness levels.


\subsection{\ournetwork{}} \label{appendix:training}
\subsubsection{Model Architecture}
\ournetwork{} uses separate encoders for each of the time series and further splits the RGB-D inputs into RGB and depth for separate encoding.
The encoder for RGB images is a pre-trained ResNet50 followed by a two-layer MLP.
The encoder for depth images is a 4-layer convolutional neural network, followed by a two-layer MLP, where each convolutional layer has a 3$\times$3 kernel and is followed by Leaky ReLU activation.
The encoder for haptics $F_t$ is a two-layer MLP 
and the encoder for end-effector poses $P_t$ is a two-layer MLP.
Each encoder outputs a vector in $\mathbb{R}^{128}$.
The four vectors are concatenated into a unified multimodal representation and then passed to an LSTM with 2 layers and a hidden size of 512.
A three-layer MLP takes output from the LSTM and produces the final output $\psi_t$. 

\subsubsection{Training}
\ournetwork{}($\sim$14M parameters) is trained using cross-entropy loss with the hyperparameters listed in Table~\ref{tab:train_hyperpara}. Training is conducted on an NVIDIA RTX 4090 GPU and completes in approximately 40 minutes.

\begin{table}[H]
\centering
\small
\caption{\textbf{Training hyperparameters for \ournetwork{}} } 
\begin{tabular}{cc}
\toprule
\textbf{Hyperparameter} & \textbf{Value} \\ \midrule
Epochs          & 200  \\
Learning rate  & 1e-3  \\
Optimizer      & Adam  \\
Batch size     & 16   \\ 
\bottomrule
\end{tabular}
\label{tab:train_hyperpara}
\end{table}

\subsubsection{Tool Calibration}
Given the utensil and skill library, tool calibration is performed once and only needs to be repeated if the tool is modified. Before deployment, we conduct tool calibration by evaluating each skill five times using the current utensil on five food items with diverse physical properties. During this process, we record each item’s physical properties and execution outcomes in natural language. The selected calibration items are raw carrot, cooked carrot, soft tofu, nuts, and cheese. The entire calibration process takes approximately 20 minutes.

\subsection{Prompting Details} \label{appendix:prompting}

\subsubsection{Perception}
We prompt GPT-4V to generate a set of candidate labels, which are then used by open-set object detectors Grounded SAM~\cite{ren2024grounded} to generate masks for each food item. The prompt we use for this application is:

\begin{lstlisting}[style = LLMQuery]
For the given image, please list the food items on the plate in a Python list format.
Here are three examples:
Example Image 1; Answer: ['chicken','broccoli','sausage']
Example Image 2; Answer: ['steak','mushroom']
Example Image 3; Answer: ['carrot','watermelon','strawberries']
<Given Image>, please list down all the food items in the plate. Follow this format: Answer: ['first_food', 'second_food', ..., 'last_food']
\end{lstlisting}

\subsubsection{Calibration}
We evaluate the utensil by executing different skills on a small set of diverse food items. During offline calibration, various utensils interact with a range of foods to assess their functional capabilities. We collect skill execution outcomes, annotated with food type and physical properties. The tool affordances are represented in natural language and later used as input to the VLM-based planner. An example of the calibration summary for the plastic fork is provided below:
\begin{lstlisting}[style = LLMQuery]
The robot interacts with various food items using a plastic fork. We summarize the history as follows:
Food Item: Nuts
Shape: Oval, Size: Bite-sized, Softness: 1, Moisture: 1, Viscosity: 2
Skill with Success Rate: Skewer 0/5, Scoop 3/5, Cut 0/5, Push 5/5, Dip 5/5

Food Item: Cheese
Shape: Block, Size: Bite-sized, Softness: 3, Moisture: 2, Viscosity: 4
Skill with Success Rate: Skewer 5/5, Scoop 3/5, Cut 5/5, Push 5/5, Dip 5/5

Food Item: Raw Carrot
Shape: Cylindrical, Size: Bite-sized, Softness: 2, Moisture: 2, Viscosity: 1
Skill with Success Rate: Skewer 0/5, Scoop 3/5, Cut 0/5, Push 5/5, Dip 5/5

Food Item: Cooked Carrot
Shape: Cylindrical, Size: Bite-sized, Softness: 2, Moisture: 3, Viscosity: 1
Skill with Success Rate: Skewer 5/5, Scoop 3/5, Cut 4/5, Push 5/5, Dip 5/5

Food Item: Soft Tofu
Shape: Cubic, Size: Large, Softness: 4, Moisture: 3, Viscosity: 2
Skill with Success Rate: Skewer 1/5, Scoop 4/5, Cut 5/5, Push 5/5, Dip 5/5

\end{lstlisting}

\subsubsection{State Estimation}
We prompt GPT-4V to estimate food physical properties based solely on visual cues. Specifically, we initialize the food property estimate using only an RGB image as input and ask the VLM to infer physical properties including shape, size, softness, moisture, and viscosity.

\begin{lstlisting}[style = LLMQuery]
<Image on the target food item>
This is a plate of <food item>.
Please estimate the physical properties of the food item, including Shape, Size, Softness, Moisture, and Viscosity, based on commonsense reasoning. For Softness, Moisture, and Viscosity, provide a score ranging from 0 to 5, similar to a 5-point Likert scale (e.g., a softness score of 1 indicates very hard, while 5 indicates very soft).

Always follow this format:
Answer: Shape: <shape> ; Size: <size>; Softness: <softness score>; Moisture: <moisture score>; Viscosity: <viscosity score> 
\end{lstlisting}

\subsubsection{Skill Selection}
We design prompting templates for skill selection. Each prompt includes a calibration summary, the history of past attempts, the available skills from the skill library, and the physical properties of the target food item. The prompt then asks the VLM to choose the most appropriate skill based on this context. We use a few-shot prompting setup with GPT-4V.
\begin{lstlisting}[style = LLMQuery]
< Calibration Summary >
The robot is using a plastic fork to pick up the food. Please select an appropriate skill by considering the food's category, shape, size, softness, moisture, and viscosity.

We briefly describe the skills as follows:
< Skill description >

The attempt history is summarized as follows:
Steak:
shape: round
size: bite-sized
softness: 2
moisture:2
viscosity: 1
scoop: success

Example Prompt 1: <Image on sausage>
This is a food item: Sausage Slice.
The robot uses a plastic fork to try picking up the food.
The estimated food physical properties are as follows. The scores range from 0 to 5, similar to a 5-pt Likert scale. For example, a softness score of 1 indicates very hard, while a score of 5 indicates very soft.
Shape: cylinder
Size: bite-sized
Softness: 3
Moisture: 2
Viscosity: 1
      
Please select an action from ['skewer, 'scoop', 'twirl', 'dip'] to pick up the food item or 
select one action from ['cut', 'push'] to manipulate items to facilitate subsequent 
acquisition. Always follow the format:  Reasoning: <your reason>. Answer: <your answer>.

Example Answer 1: 
Reasoning: Sausage slices are moderately firm to maintain their structure. Skewering is suitable as the fork can easily pierce them without breaking them apart.
Answer: skewer

Example Prompt 2: <Image on mashed potatoes>
This is a food item: Mashed Potatoes.
The robot uses a plastic fork to try picking up the food.
The estimated food physical properties are as follows. The scores range from 0 to 5, similar to a 5-pt Likert scale. For example, a softness score of 1 indicates very hard, while a score of 5 indicates very soft.
Shape: Amorphous
Size: bite-sized
Softness: 4
Moisture: 3
Viscosity: 2
      
Please select an action from ['skewer, 'scoop', 'twirl', 'dip'] to pick up the food item or 
select one action from ['cut', 'push'] to manipulate items to facilitate subsequent 
acquisition. Always follow the format:  Reasoning: <your reason>. Answer: <your answer>.

Example Answer 2: 
Reasoning: The food is soft and moist, making it suitable to scoop rather than skewer or cut. The viscosity indicates it will adhere moderately to the fork.
Answer: scoop

This is a food item: Tofu. <image>
The robot uses a plastic fork to try picking up the food.
Food Item: Tofu
Shape: Cubic
Size: bite-sized
Softness: 4
Moisture: 3
Viscosity: 2


Please select an action from ['skewer, 'scoop', 'twirl', 'dip'] to pick up the food item or 
select one action from ['cut', 'push'] to manipulate items to facilitate subsequent 
acquisition. Always follow the format:  Reasoning: <your reason>. Answer: <your answer>.


\end{lstlisting}

\end{document}